\documentclass[11pt]{article}
\usepackage{authblk}

\usepackage{times}
\usepackage{url}
\usepackage{latexsym}
\usepackage{latexsym}
\usepackage{graphicx}
\usepackage{caption}
\usepackage{makecell}
\usepackage{amssymb}
\usepackage{booktabs}
\usepackage{parskip}
\usepackage{tabularx}
\usepackage{verbatim}
\usepackage{subfig}
\usepackage{hyperref}
\usepackage{coling2018}
\usepackage{soul}
\usepackage[usenames, dvipsnames]{color}
\usepackage{wrapfig}

\title{Power Networks: A Novel Neural Architecture to Predict Power Relations}

\author{\textbf{Michelle Lam}*}
\author{\textbf{Catherina Xu}*}
\author{\textbf{Angela Kong}}
\author{\textbf{Vinodkumar Prabhakaran}}
\affil{Department of Computer Science, Stanford University, Stanford, CA, USA.}
\affil[ ]{\tt {\{mlam4, yuex, akong2, vinod\}@cs.stanford.edu}}

\begin{document}

\maketitle


\blfootnote{
    * Authors (listed in alphabetical order) contributed equally.
}

\begin{abstract}
Can language analysis reveal the underlying social power relations that exist between participants of an interaction? Prior work within NLP has shown promise in this area, but the performance of automatically predicting power relations using NLP analysis of social interactions remains wanting. In this paper, we present a novel neural architecture that captures manifestations of power within individual emails which are then aggregated in an order-preserving way in order to infer the direction of power between pairs of participants in an email thread. We obtain an accuracy of 80.4\%, a 10.1\% improvement over state-of-the-art methods, in this task. We further apply our model to the task of predicting power relations between individuals based on the entire set of messages exchanged between them; here also, our model significantly outperforms the 70.0\% accuracy using prior state-of-the-art techniques, obtaining an accuracy of 83.0\%.
\end{abstract}

\section{Introduction}
\label{intro}
With the availability and abundance of linguistic data that captures different avenues of human social interactions, there is an unprecedented opportunity to expand NLP to not only understand language, but also to understand the \textit{people} who speak it and the \textit{social relations} between them. Social power structures are ubiquitous in human interactions, and since power is often reflected through language, computational research at the intersection of language and power has gained interest recently.
This research has been applied to a wide array of domains such as Wikipedia talk pages \cite{Strzalkowski:2010,Taylor2012,Danescu2012,SwayamdiptaAndRambow2012}, blogs \cite{rosenthal2014detecting} as well as workplace interactions \cite{bramsen,gilbert,prabhakaran_thesis}.


The corporate environment is one social context in which power dynamics have a clearly defined structure and shape the interactions between individuals, making it an interesting case study on how language and power interact. 
Organizations stand to benefit greatly from being able to detect power dynamics within their internal interactions, in order to address disparities and ensure inclusive and productive workplaces. 
For instance, \cite{cortina2001incivility} reports that women are more likely to experience incivility, often from superiors. It has also been shown that incivility may breed more incivility \cite{harold2015effects}, and that it can lead to increased stress and lack of commitment \cite{miner2012experiencing}.

Prior work has investigated the use of NLP techniques to study manifestations of different types of power using the Enron email corpus \cite{diesner2005communication,prabhakaran-rambow-diab:2012:PAPERS,prabhakaran-rambow:2013:IJCNLP,prabhakaran}. While early work \cite{bramsen,gilbert} focused on surface level lexical features aggregated at corpus level, more recent work has looked into the thread structure of emails as well \cite{prabhakaran}. However, both \cite{bramsen,gilbert} and \cite{prabhakaran} group all messages sent by an individual to another individual (at the corpus-level and at the thread-level, respectively) and rely on word-ngram based features extracted from this concatenated text to infer power relations. They ignore the fact that the text comes from separate emails, and that there is a sequential order to them. 

In this paper, we propose a hierarchical deep learning architecture for power prediction, using a combination of Convolutional Neural Networks (CNN) to capture linguistic manifestations of power in individual emails, and a Long Short-Term Memory (LSTM) that aggregates their outputs in an order-preserving fashion. We obtain significant improvements in accuracy on the corpus-level task (82.4\% over 70.0\%) and on the thread-level task (80.4\% over 73.0\%) over prior state-of-the-art techniques.


%
%
\blfootnote{
    %
    %
    %
    %
    %
    %
    \hspace{-0.65cm}  
    This work is licensed under a Creative Commons 
    Attribution 4.0 International License.
    License details:
    \url{http://creativecommons.org/licenses/by/4.0/}
}

\section{Data and Problem Formulation}

We use the version of the Enron Email corpus released by \newcite{agarwal}
that captures the organizational power relation between 13,724 pairs of Enron employees, in addition to the reconstructed thread structure of email messages added by \newcite{yeh2006email}. 
We mask greetings and signature lines in the email content to prevent our model from being biased by the roles held by specific employees.


\begin{table}[h]
    \centering
    \label{table_data_stats}
    \begin{tabular}{ *5c }
    \midrule
    Entity type & \# of Pairs  & Train  & Dev  & Test  \\ 
    \midrule
    Per-Thread & 15,048 & 7,510 & 3,578 & 3,960\\
    Grouped & 3,755 & 2,253 & 751 & 751\\
    \midrule
    \end{tabular}
    \caption{ Data instance statistics by problem formulation.}
\end{table}

Prior work on NLP approaches to predict power in organizational email has used two different problem formulations --- \textit{Per-Thread} and \textit{Grouped}. We investigate both formulations in this paper. Table 1 shows the number of data instances in each problem formulation.

\textbf{Per-Thread}: This formulation was introduced by \newcite{prabhakaran} in which, for a given thread \textit{t} and a pair of related interacting participant pairs (\textit{A}, \textit{B}), the direction of power between \textit{A} and \textit{B} is predicted (where the assignment of labels \textit{A} and \textit{B} is arbitrary). The participants in these pairs are 1) \textit{interacting}: at least one message exists in the thread such that either \textit{A} is the sender and \textit{B} is a recipient or vice versa, and 2) \textit{related}: \textit{A} and \textit{B} are related by a dominance relation (either superior or subordinate) based on the organizational hierarchy. As in \cite{prabhakaran}, we exclude pairs of employees who are peers, and we use the same train-dev-test splits so our results are comparable.

\textbf{Grouped}: Here, we group all emails \textit{A} sent to \textit{B} across all threads in the corpus, and vice versa, and use these sets of emails to predict the power relation between \textit{A} and \textit{B}. This formulation is similar those in \cite{bramsen,gilbert}, but our results are not directly comparable since, unlike them, we rely on the ground truth of power relations from \cite{agarwal}; however, we created an SVM model that uses word-ngram features similar to theirs as a baseline to our proposed neural architectures.


\section{Methods}

The inputs to our models take on two forms: 
\textbf{Lexical features:} We represent each email as a series of tokenized words, each of which is represented by a 100-dimensional GloVe vector pre-trained on Wikipedia and Gigaword \cite{glove}. We cap the email length at a maximum of 200 words. 
\textbf{Non-lexical features:} We incorporate the structural non-lexical features identified as significant by \newcite{prabhakaran} for the \textit{Grouped} problem formulation. We used (1) the average number of recipients and (2) the average number of words in each email for each individual; these features were concatenated into a single input vector. We investigate the following three network architectures, in increasing order of complexity, to train our model:

\begin{figure}[h]
    \centering
    \small
    \includegraphics[width=0.65\textwidth]{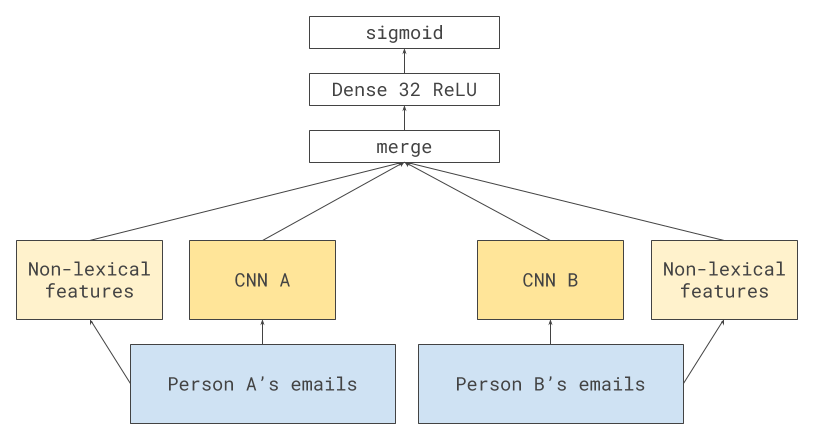}
    \caption{Batched emails (Batched-CNN)}
    \label{fig:approach1}
\end{figure}
\noindent \textbf{Approach 1: Batched emails (Batched-CNN).} 
In this model (see Figure 1), all of \textit{A}'s emails to \textit{B} are batched and fed into a Convolutional Neural Network (CNN), and the same operation is performed for \textit{B}'s emails to \textit{A}. The format of this input is described earlier in this section. 
This representation can be thought of as a neural equivalent of the SVM-based approaches in prior work, since they merge together all emails in either direction as a single unit.
Then, the output of these two CNNs is merged with the non-lexical features from \textit{A}'s emails and \textit{B}'s emails, passed through a dense layer with rectified linear unit (ReLU) activation, and fed to a sigmoid classifier that predicts the power relation between \textit{A} and \textit{B}. 


\begin{figure}[h]
    \centering
    \small
    \includegraphics[width=0.95\textwidth]{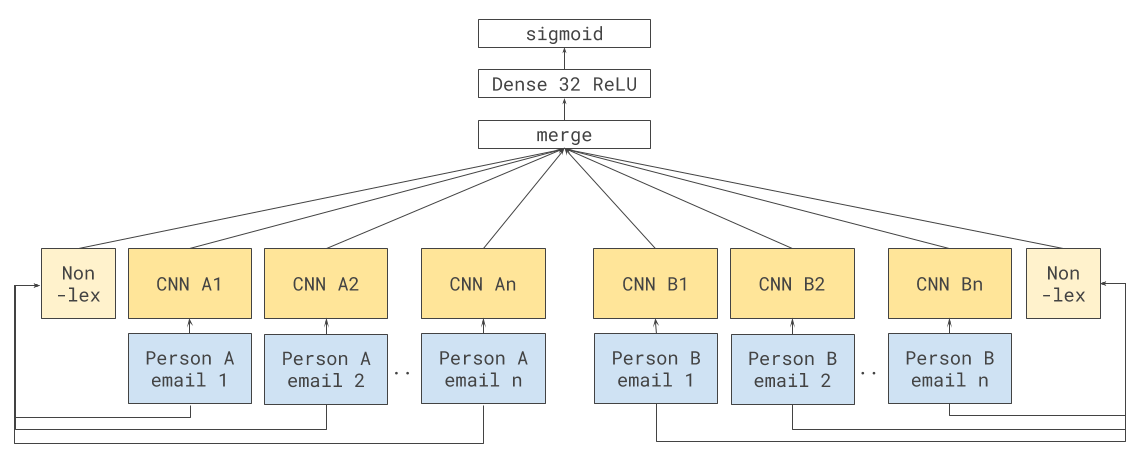}
    \label{fig:approach2}
    \caption{Separated emails (Separated-CNN)}
\end{figure}
\noindent \textbf{Approach 2: Separated emails (Separated-CNN).} 
In this model (see Figure 2), we capture the essence of individual emails by separating them in the model input. As in Batched-CNN, we separate \textit{A}'s and \textit{B}'s emails, but here we feed \textit{each} email as input to a CNN. 
The motivation here is to first capture local patterns from individual emails.
We then merge the output of these CNNs with the non-lexical features from \textit{A}'s and \textit{B}'s emails, pass this to a dense layer with ReLU activation, and pass the result to a sigmoid classifier that predicts the power relation.

\begin{figure}[h]
    \centering
    \small
    \vspace{-3mm}
    \includegraphics[width=0.75\textwidth]{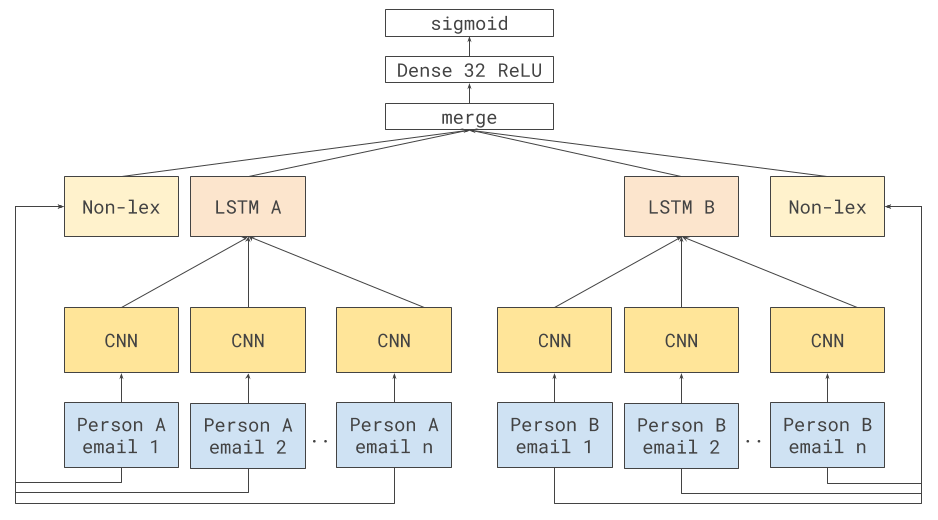}
    \caption{Sequential emails (Sequential-CNN-LSTM)} \label{fig:approach3}
\end{figure}
\noindent \textbf{Approach 3: Sequential emails (Sequential-CNN-LSTM).} Finally, we use a third model where we account for the temporal order of emails, which may be important in the case of the \textit{Per-Thread} formulation.
In this model (see \autoref{fig:approach3}), we separate each individual's emails, feed each email to a CNN, and pass the sequence of CNN outputs for each email to a Long Short-Term Memory network (LSTM) for that individual. We then merge the resulting output of the two LSTMs with the non-lexical features from each individual's emails, pass it on to a dense layer with ReLU activation, and then to a sigmoid classifier for the final prediction.

\section{Experiments and Results}
\label{experiments}

We use support vector machine (SVM) based approaches as our baseline, since they are the state-of-the art in this problem \cite{prabhakaran,bramsen,gilbert}.
We use the performance reported by \cite{prabhakaran} using SVM as baseline for the \textit{Per-Thread} formulation (using the same train-dev-test splits) and implemented an SVM baseline for the \textit{Grouped} formulation (not directly comparable to performance reported by \cite{bramsen,gilbert}).

For each of our neural net models, we trained for 30-70 epochs until 
the performance on the development set stopped improving, in order
to avoid overfitting. We used Hyperas to tune hyperparameters on our development dataset for the same set of parameter options for each task formulation, varying activation functions, hidden layer size, batch size, dropout, number of filters, and number of words to include per email.\footnote{\url{https://github.com/maxpumperla/hyperas}}

\begin{wraptable}{r}{0.52\linewidth}
\vspace{-2mm}
\centering
\label{my-label}
\begin{tabularx}{0.52\textwidth}{l *{2}{c}}
\toprule
Model & Per-Thread & Grouped \\ \midrule
SVM Baseline & 73.0 & 70.0       \\ \midrule
Batched-CNN & 78.7 &  82.0      \\
Separated-CNN     & 79.8 & \textbf{83.0} \\ 
Sequential-CNN-LSTM &  \textbf{80.4}  & 82.4 \\ \bottomrule
\end{tabularx}
\caption{Accuracy obtained using different models\label{tab_results}}
SVM Baseline: \cite{prabhakaran} 
\end{wraptable}
Table~\ref{tab_results} presents the accuracy obtained using different models. All of our models significantly outperformed the SVM baseline in both task formulations. In the \textit{Per-Thread} formulation, we obtained the best accuracy of 80.4\% using the Sequential-CNN-LSTM approach, compared to the 73.0\% reported by \cite{prabhakaran}. This is also a marked improvement over the simpler Batched-CNN and Separated-CNN models. This suggests that both temporal and local email features aid in the power prediction task within the \textit{Per-Thread} formulation.
In the \textit{Grouped} formulation, the Separated-CNN model obtained the best accuracy of 83.0\%, outperforming the Sequential-CNN-LSTM accuracy of 82.4\%. 
We hypothesize that this is because the grouped formulation does not inherently have a temporal structure between emails, unlike the thread formulation where Sequential-CNN-LSTM is able to tap into the temporal structure.

\autoref{table:qualitative-results} presents
a few emails from our corpus, along with the true and predicted labels for the power relation between their sender and recipient(s). 
Our model seems to pick up on linguistic signals of lack of power such as relinquishing agency (\textit{let me know who you'd like us to work with}), and status reports (\textit{model is nearly completed}), as well as overt displays of power such as \textit{I personally would like to see the results} and \textit{we need to end all payments}. On the other hand, the model picks up on the phrasing in \textit{don't use the ftp site} as displaying superiority while the superiority displayed here may have been derived from the expertise the subordinate has in file-transfer protocols. Similarly, the model may have misunderstood the overtly polite phrasings in the last email sent by a superior to be subordinate-like behavior.  
This sheds light on an important challenge in this task: superiors don't express their superiority in all emails, and subordinates may sometimes display power derived from other sources such as expertise. In such cases where text features alone are not informative enough, signals from additional non-lexical features may be key to accurate classification.


\begin{table}[ht]
\small
\centering 
\begin{tabular*}{1.0\linewidth}{p{0.7\linewidth} p{0.12\linewidth} p{0.12\linewidth}}
\toprule
\textbf{Text extracted from email} & \textbf{Actual} & 
\textbf{Predicted} \\ 
\midrule
\setulcolor{Green}\ul{Let me know who you'd like us to work with} in your group. The Adaytum planning model is \ul{nearly completed}. & \color{Black}{Subordinate} & \color{ForestGreen}{Subordinate} \\
\midrule
Vince is hosting on Wharton and a project they are doing for us, \setulcolor{Green}\ul{I personally would like to see the results of that} before doing more with Wharton. & \color{Black}{Superior} & \color{ForestGreen}{Superior} \\
\midrule
\setulcolor{Green}\ul{We need to end all payments} as of December 31, 2001. & \color{Black}{Superior} & \color{ForestGreen}{Superior} \\
\midrule
\setulcolor{Green}\setulcolor{Red}\ul{Don't use the ftp site} while I am away [...] I will check my messages \ul{when I return}. & \color{Black}{Subordinate} & \color{Red}{Superior} \\
\midrule
Here is the draft letter \setulcolor{Red}\ul{for your consideration}. \ul{Please} distribute this to the members of the legal team at Enron. \ul{Thank you for your assistance}, and have a \ul{happy holiday season.} & \color{Black}{Superior} & \color{Red}{Subordinate} \\
\bottomrule
\end{tabular*}
\caption{Example power labels from Separated-CNN on the Grouped formulation. For \textit{correct} labels, text segments that may signal the power relation are \setulcolor{Green}\ul{underlined in green}; for \textit{incorrect} labels, potentially confusing power signals are \setulcolor{Red}\ul{underlined in red}. (text segments chosen based on our qualitative judgment).} 
\label{table:qualitative-results}
\end{table}

\section{Concluding Discussions}

In this paper, we investigated the intersection between language and power in the corporate domain via neural architectures grounded in an understanding of how expressions of power unfold in email. Our Sequential-CNN-LSTM model, which utilizes an LSTM to capture the temporal relations underlying per-email features, achieved 80.4\% accuracy in predicting the direction of power between participant pairs in individual email threads, which is a 10.1\% accuracy improvement over the state-of-the-art approach \cite{prabhakaran}. 
Our Separated-CNN model also obtains an accuracy of 83.0\%  in predicting power relations between individuals based on the entire set of messages exchanged between them,
a significant boost over 70.0\% accuracy obtained using traditional methods. We also present a qualitative error analysis that sheds light on the patterns that confuse the model.


To further our work, we plan to granularize the level at which features are learned. We hypothesize that by training a CNN on each \textit{sentence} rather than email, the model will better capture mid-level indicators of power that occur between the word level and email level. We will also investigate ways to better incorporate structural features by accounting for their relevance to a holistic judgment of power; for example, features like gender and temporal position in a thread are more suited to merge with a higher level of the architecture like the per-individual LSTMs while features like number of email tokens are more suited to merge at the low level of the per-email CNNs. Lastly, we plan to incorporate additional datasets such as the Avocado Research Email Collection \cite{oard2015avocado} to study cross-corpora performance. 

\section{Acknowledgements}
We would like to thank Christopher Manning, Richard Socher, and Arun Chaganty, who served as course instructors for CS 224N (Natural Language Processing with Deep Learning, the course in which M. Lam, C. Xu and A. Kong developed the first iteration of this work) for their feedback on the project. V. Prabhakaran was supported by a John D. and Catherine T. MacArthur Foundation award granted to Jennifer Eberhardt and Dan Jurafsky. We also thank anonymous reviewers for their useful feedback. 

\bibliography{references}

\begin{thebibliography}{}

\bibitem[\protect\citename{Agarwal \bgroup et al.\egroup }2012]{agarwal}
Apoorv Agarwal, Adinoyi Omuya, Aaron Harnly, and Owen Rambow.
\newblock 2012.
\newblock A Comprehensive Gold Standard for the Enron Organizational Hierarchy.
\newblock In {\em Proceedings of the 50th Annual Meeting of the Association for
  Computational Linguistics (Volume 2: Short Papers)}, pages 161--165, Jeju
  Island, Korea, July. ACL.

\bibitem[\protect\citename{Bramsen \bgroup et al.\egroup }2011]{bramsen}
Philip Bramsen, Martha Escobar-Molano, Ami Patel, and Rafael Alonso.
\newblock 2011.
\newblock Extracting Social Power Relationships from Natural Language.
\newblock In {\em Proceedings of the 49th Annual Meeting of the Association for
  Computational Linguistics: Human Language Technologies}, pages 773--782,
  Portland, Oregon, USA, June. ACL.

\bibitem[\protect\citename{Cortina \bgroup et al.\egroup
  }2001]{cortina2001incivility}
Lilia~M Cortina, Vicki~J Magley, Jill~Hunter Williams, and Regina~Day Langhout.
\newblock 2001.
\newblock Incivility in the Workplace: Incidence and Impact.
\newblock {\em Journal of Occupational Health Psychology}, 6(1):64.

\bibitem[\protect\citename{Danescu-Niculescu-Mizil \bgroup et al.\egroup
  }2012]{Danescu2012}
Cristian Danescu-Niculescu-Mizil, Lillian Lee, Bo~Pang, and Jon Kleinberg.
\newblock 2012.
\newblock Echoes of Power: Language Effects and Power Differences in Social
  Interaction.
\newblock In {\em Proceedings of the 21st International Conference on World
  Wide Web}, WWW '12, New York, NY, USA. ACM.

\bibitem[\protect\citename{Diesner \bgroup et al.\egroup
  }2005]{diesner2005communication}
Jana Diesner, Terrill~L Frantz, and Kathleen~M Carley.
\newblock 2005.
\newblock Communication Networks from the Enron Email Corpus “It's Always
  About the People. Enron is no Different”.
\newblock {\em Computational \& Mathematical Organization Theory},
  11(3):201--228.

\bibitem[\protect\citename{Gilbert}2012]{gilbert}
Eric Gilbert.
\newblock 2012.
\newblock Phrases That Signal Workplace Hierarchy.
\newblock {\em CSCW '12 Proceedings of the ACM 2012 conference on Computer
  Supported Cooperative Work}, pages 1037--1046.

\bibitem[\protect\citename{Harold and Holtz}2015]{harold2015effects}
Crystal~M Harold and Brian~C Holtz.
\newblock 2015.
\newblock The Effects of Passive Leadership on Workplace Incivility.
\newblock {\em Journal of Organizational Behavior}, 36(1):16--38.

\bibitem[\protect\citename{Miner \bgroup et al.\egroup
  }2012]{miner2012experiencing}
Kathi~N Miner, Isis~H Settles, Jennifer~S Pratt-Hyatt, and Christopher~C Brady.
\newblock 2012.
\newblock Experiencing Incivility in Organizations: The Buffering Effects of
  Emotional and Organizational Support.
\newblock {\em Journal of Applied Social Psychology}, 42(2):340--372.

\bibitem[\protect\citename{Oard \bgroup et al.\egroup }2015]{oard2015avocado}
Douglas Oard, William Webber, David Kirsch, and Sergey Golitsynskiy.
\newblock 2015.
\newblock Avocado Research Email Collection.
\newblock {\em Philadelphia: Linguistic Data Consortium}.

\bibitem[\protect\citename{Pennington \bgroup et al.\egroup }2014]{glove}
Jeffrey Pennington, Richard Socher, and Christopher~D. Manning.
\newblock 2014.
\newblock GloVe: Global Vectors for Word Representation.
\newblock {\em Proceedings of the 2014 Conference on Empirical Methods in
  Natural Language Processing (EMNLP)}, pages 1532--1543.

\bibitem[\protect\citename{Prabhakaran and
  Rambow}2013]{prabhakaran-rambow:2013:IJCNLP}
Vinodkumar Prabhakaran and Owen Rambow.
\newblock 2013.
\newblock Written Dialog and Social Power: Manifestations of Different Types of
  Power in Dialog Behavior.
\newblock In {\em Proceedings of the Sixth International Joint Conference on
  Natural Language Processing}, pages 216--224, Nagoya, Japan, October. Asian
  Federation of NLP.

\bibitem[\protect\citename{Prabhakaran and Rambow}2014]{prabhakaran}
Vinodkumar Prabhakaran and Owen Rambow.
\newblock 2014.
\newblock Predicting Power Relations between Participants in Written Dialog
  from a Single Thread.
\newblock {\em Proceedings of the 52nd Annual Meeting of the Association for
  Computational Linguistics}, pages 339--344.

\bibitem[\protect\citename{Prabhakaran \bgroup et al.\egroup
  }2012]{prabhakaran-rambow-diab:2012:PAPERS}
Vinodkumar Prabhakaran, Owen Rambow, and Mona Diab.
\newblock 2012.
\newblock Who's (Really) the Boss? Perception of Situational Power in Written
  Interactions.
\newblock In {\em Proceedings of COLING 2012}, pages 2259--2274, Mumbai, India,
  December. The COLING 2012 Organizing Committee.

\bibitem[\protect\citename{Prabhakaran}2015]{prabhakaran_thesis}
Vinodkumar Prabhakaran.
\newblock 2015.
\newblock {\em Social Power in Interactions: Computational Analysis and
  Detection of Power Relations}.
\newblock {Ph.D.} thesis, Columbia University.

\bibitem[\protect\citename{Rosenthal}2014]{rosenthal2014detecting}
Sara Rosenthal.
\newblock 2014.
\newblock Detecting Influencers in Social Media Discussions.
\newblock {\em XRDS: Crossroads, The ACM Magazine for Students}, 21(1):40--45.

\bibitem[\protect\citename{Strzalkowski \bgroup et al.\egroup
  }2010]{Strzalkowski:2010}
Tomek Strzalkowski, George~Aaron Broadwell, Jennifer Stromer-Galley, Samira
  Shaikh, Sarah Taylor, and Nick Webb.
\newblock 2010.
\newblock Modeling Socio-Cultural Phenomena in Discourse.
\newblock In {\em Proceedings of the 23rd International Conference on COLING
  2010}, Beijing, China, August. Coling 2010 Organizing Committee.

\bibitem[\protect\citename{Swayamdipta and
  Rambow}2012]{SwayamdiptaAndRambow2012}
Swabha Swayamdipta and Owen Rambow.
\newblock 2012.
\newblock The Pursuit of Power and Its Manifestation in Written Dialog.
\newblock {\em 2012 IEEE Sixth International Conference on Semantic Computing},
  0:22--29.

\bibitem[\protect\citename{Taylor \bgroup et al.\egroup }2012]{Taylor2012}
Sarah~M. Taylor, Ting Liu, Samira Shaikh, Tomek Strzalkowski, George~Aaron
  Broadwell, Jennifer Stromer-Galley, Umit Boz, Xiaoai Ren, Jingsi Wu, and
  Feifei Zhang.
\newblock 2012.
\newblock {Chinese and American Leadership Characteristics: Discovery and
  Comparison in Multi-party On-Line Dialogues}.
\newblock In {\em ICSC}, pages 17--21.

\bibitem[\protect\citename{Yeh and Harnly}2006]{yeh2006email}
J.Y. Yeh and A.~Harnly.
\newblock 2006.
\newblock {Email Thread Reassembly Using Similarity Matching}.
\newblock In {\em Third Conference on Email and Anti-Spam (CEAS)}, pages
  27--28.

\end{thebibliography}
\bibliographystyle{acl}
\clearpage

\end{document}